\RequirePackage{fix-cm}
\documentclass[smallextended]{svjour3}       
\smartqed  
\usepackage{graphicx}
\journalname{Earth Science Informatics}
\usepackage{epsf}
\usepackage{graphicx}
\usepackage{float}    
\def\etal{{\it et al.}}

\def\ddt#1{{{\partial #1}\over {\partial t}}} 
 
\def\ddy#1{{{\partial #1}\over {\partial y}}} 

\def\ddx#1{{{\partial #1}\over {\partial x}}}

\begin{document}
\title{ A spectral optical flow method for determining velocities from
digital imagery}
\author{Neal Hurlburt \& Steve Jaffey} 
\institute{Lockheed Martin Solar and Astrophysics Laboratory, \ A021S,
Building 252, 3251 Hanover Street, Palo Alto CA 94304}
\date{Received: date / Accepted: date}

\maketitle

\begin{abstract} We present a method for determining surface flows from solar
images based upon optical flow techniques. We apply the method to sets of images obtained by a variety of solar imagers to assess its performance. The {\tt opflow3d} procedure is shown to extract accurate velocity estimates when provided perfect test data and quickly generates results consistent with completely distinct methods when applied on global scales. We also validate it in detail by comparing it to an established method when applied to high-resolution datasets and find that it provides comparable results without the need to tune, filter or otherwise preprocess the images before its application. 
\end{abstract}

\section{Introduction}

The most powerful events observed in the solar system are the result of convective dynamics in the outer layers of the Sun. 
Researchers trying to understand these dynamics need tools to measure and study these turbulent motions and their interaction 
with the local magnetic field.  Several methods for deducing the flows in the solar
photosphere (the visible surface of the sun) have been developed over
the years. Many of these have been based upon some form of
local correlation tracking (LCT) \cite{TTS86,N87} where pairs of
successive images of the photosphere are broken into sub-images which
are then shifted relative to each other to find an optimal relative
shift. This shift is then associated with the local mean velocity of
the flow. These methods appear to work well on high-resolution images collected from both
ground-based  \cite{B_al88}  and space-based \cite{S_al88,TTS89} observatories. There have been a few attempts to assess and compare the performance of such methods over the past twenty years including  those of Hurlburt \etal ~\cite{H95}, Welsch \etal \cite{W07}, Chae and Sakurai \cite{CS2008} and most recently, Verma, Steffen and Denker \cite{Verma13}.  These have used a variety of data input for their assessments, from those derived from simulations of MHD flows in the photosphere, to creating image sets from solar images by applying known distortions, to direct comparisons with real solar data. All tested methods gave consistent results. However the best method was somewhat dependent on the test and the choice of the respective parameters for each method.

Alternative methods have been developed in other fields with similar goals.  Machine vision researchers developed optical flow methods for deducing the relative motions of objects in digital images \cite{B2011,S2014} and atmospheric scientists developed various methods for deducing winds on Earth from cloud motions in satellite images \cite{L2013,C1999}. Experimentalists in other branches of fluid dynamics, including blood flow measurements deduced from x-ray images \cite{N2012} and flows with various tracer particles in suspensions \cite{W2001}, have explored similar methods.  One difference between the typical machine vision problem and that of deducing fluid velocities is that the former seek motions of discrete objects while the latter seeks motions of continuous flows. 
Applying optical flow methods such as those described by J\"ahne \cite{J93} on high-resolution solar images typically underestimate the velocity of the flows. In part this is due to the simple spatial averaging used in deriving the velocity. 

Hurlburt \cite{H99} presented a method which does not make use of such spatial
averaging. Instead the flows are assumed to be smooth and continuous -- being 
represented by a truncated  Fourier series.  Here  we present a detailed description of the method and assess it using previously
developed tests and compare it to local correlation tracking methods. 

\section{Method}

The basic assumption is that the visible pattern observed in the
fluid, as measured by the local intensity $I,$ will be advected by
the velocity field ${\bf v}$ and hence should satisfy the
equation
\begin{equation}
\ddt{I}+ {\bf v} \cdot \nabla I = 0 \label{eq:advect}
\end{equation}
In the case of solar physics the pattern is typically formed by
convective motions in the photosphere, which are clearly visible in
white-light images and which appear to be
advected by larger scale flow fields.   Images are collected
frequently relative to the flow speeds, such that the displacement of
the pattern between any sequential pair of images is less than a
pixel. The problem is to determine ${\bf v}$ from a time sequence of
two dimensional images $I(x,y,t)$ in the presence of measurement noise and other ``noise'' sources, such as the
acoustic oscillations present in the solar atmosphere or missing frames due to data dropouts. Using equation
(\ref{eq:advect}) we can seek the best fit velocity field ${\bf v}_f$
using least squares. First we form the merit function of the fit for
the full dataset $I(x,y,t)$.
\begin{equation}
\chi({\bf v}_f)^2 = \Sigma_t\Sigma_x\Sigma_y \bigl(\ddt{I}+ {\bf v}_f \cdot \nabla I\bigr)^2 \label{eq:merit}
\end{equation}
which we seek to minimize. Here the sums are taken over the discrete pixel and frame coordinates for $x,y$ and $t.$ If the velocity field ${\bf v}$ is a
continuous field, we can express the fit velocity ${\bf v_f}$ as a Fourier series
\begin{equation}
{\bf v}_f=  \Sigma_{i=-N_x}^{i=N_x}\Sigma_{i=-N_y}^{i=N_y}(\alpha_{ij}{\bf\hat x} +
\beta_{ij}{\bf\hat y}) e^{-2\pi{\it I}( ix/X + jy/Y)} \label{eq:vel}
\end{equation}

\noindent where {$\bf\hat x$} and {$\bf\hat y$} are unit vectors, $\alpha_{ij}$
and $\beta_{ij}$ are complex amplitudes and $ N_x$ and $N_y$ are the
number of Fourier modes retained in the expansion. Substituting this
into equation (\ref{eq:merit}) and differentiating with respect to
$\alpha_{kl}$ and $\beta_{kl}$ results in the system of equations
\begin{eqnarray}
\Sigma_t\Sigma_x\Sigma_y \bigl(\ddt{I}+ {\bf v}_f \cdot
\nabla I\bigr) \ddx{I} e^{-2\pi{\rm I}( kx/X + ly/Y)} &=& 0 \\
\Sigma_t\Sigma_x\Sigma_y \bigl(\ddt{I}+ {\bf v}_f \cdot \nabla I\bigr) \ddy{I}
e^{-2\pi{\rm I}( kx/X + ly/Y)} &=& 0 \label{eq:least}
\end{eqnarray}
This can be reorganized to form the complex system of equations
\begin{equation}
\Sigma_i\Sigma_j \bigl(\alpha_{ij} M^{xx}_{ijkl} + \beta_{ij}
M^{xy}_{ijkl} \bigr) = R^x_{kl}, \qquad
\Sigma_i\Sigma_j \bigl(\alpha_{ij} M^{xy}_{ijkl} + \beta_{ij} M^{yy}_{ijkl}\bigr) = R^y_{kl} \label{eq:spectral}
\end{equation}
where
\begin{eqnarray}
\nonumber R^x_{kl} &=&- \Sigma_x\Sigma_y\Sigma_t\biggl(\ddt{I}\ddx{I}\biggr)e^{-2\pi{\rm I}(
kx/X + ly/Y)}\\
\nonumber R^y_{kl} &=&- \Sigma_x\Sigma_y\Sigma_t\biggl(\ddt{I}\ddy{I}\biggr)e^{-2\pi{\rm I}(
kx/X + ly/Y)}\\
\nonumber M^{xx}_{ijkl} &=& \Sigma_x\Sigma_y\Sigma_t\biggl(\ddx{I}\biggr)^2
e^{-2\pi{\rm I}( (k+i)x/X + (l+j)y/Y)}\\
\nonumber M^{xy}_{ijkl} &=& \Sigma_x\Sigma_y\Sigma_t\biggl(\ddx{I}\ddy{I}\biggr)
e^{-2\pi{\rm I}( (k+i)x/X + (l+j)y/Y)}\\
M^{yy}_{ijkl} &=& \Sigma_x\Sigma_y\Sigma_t\biggl(\ddy{I}\biggr)^2 e^{-2\pi{\rm I}( (k+i)x/X + (l+j)y/Y)}
 \label{eq:system}
\end{eqnarray}

These matrices consist of discrete Fourier transforms of the various,
time-averaged products of the spatial and temporal derivatives of the
image.  The matrices of the complex linear system (\ref{eq:spectral}) for
the spectral amplitudes $\alpha_{ij}$, $\beta_{ij}$ can be combined to form a single hermitian matrix of size $(8N_x \times N_y)^2.$  

This method has been implemented in an IDL \footnote{Trademark,
Exelis.} routine {\tt opflow3d} and is available as part of the SolarSoft environment \cite{Freeland}. The time derivative is
evaluated using finite differencing between sequential images while
the spatial derivatives is evaluated using 4th order finite differences
on the average of the two images used for the time derivative. The
matrices are then computed for the entire time-space cube $I(x,y,t)$
and the system is solved.

Solving the system using direct
methods can quickly become expensive, scaling as $(N_x \times N_y)^3$. The method requires 70
seconds on an 2013-vintage iMac workstation  (with 3.4GHz processor and 32GB of memory) to obtain the
solutions for $N_x = N_y =24$ on a $1024 \times 1024$ image.  This is partially offset by the fact that the method requires no preprocessing or filtering and that it fits flows over many instances in time in one go. The matrices in (\ref{eq:system}) can also be reused in subsequent calculations with minimal additional expense. Performance could be further optimized by taking advantage of the matrix structure in equation (\ref{eq:system}) which has a blocked-Toeplitz-Toeplitz-block (BTTB) structure \cite{C1999}.

\section{Evaluation and comparisons}

\begin{figure}
\includegraphics[width=\columnwidth]{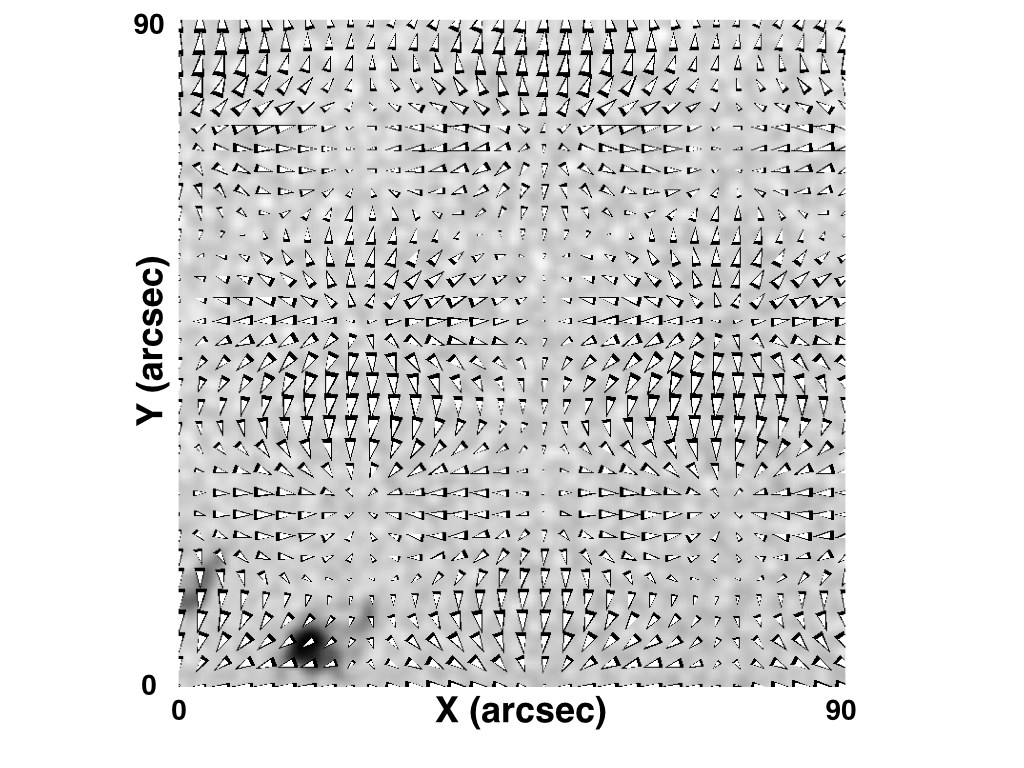}
\caption{Comparison between a derived velocity field and input field
used to distort a solar image (background). The input (derived)
velocity field is displayed with black (white) arrows whose areas are
proportional to the magnitude of the local velocity. The relative
scale of the white arrows has been reduced slightly for aid in
comparison.}
\label{fig:H95}
\end{figure}

As a first test of the method, we take the simulated observations developed by Hurlburt \etal ~\cite{H95}. Using  a sixth-order accurate numerical scheme \cite{HR00} they took a single intensity image $I_0$  of solar
granulation and evolved it with equation
(\ref{eq:advect}) with a steady velocity field. They then degraded and resampled the image to represent the expected resolution of the Michelson Doppler Imager (MDI) on the Solar and Heliospheric Observatory \cite{MDI95}. Since there is no source of noise and the imposed flows are themselves based on Fourier modes, we expect and observe that  the {\tt opflow3d} method can recover the flow with a high accuracy. The results for a case where $N_x = N_y = 4$ on a $140 \times 140$ pixel
image is displayed in figure \ref{fig:H95}, along with the known input
velocty field and a sample image. The two sets of arrows, corresponding to the known (black) and derived (white) velocity fields are almost perfectly correlated, both in direction and magnitude. 

\begin{figure}
\includegraphics[width=\columnwidth]{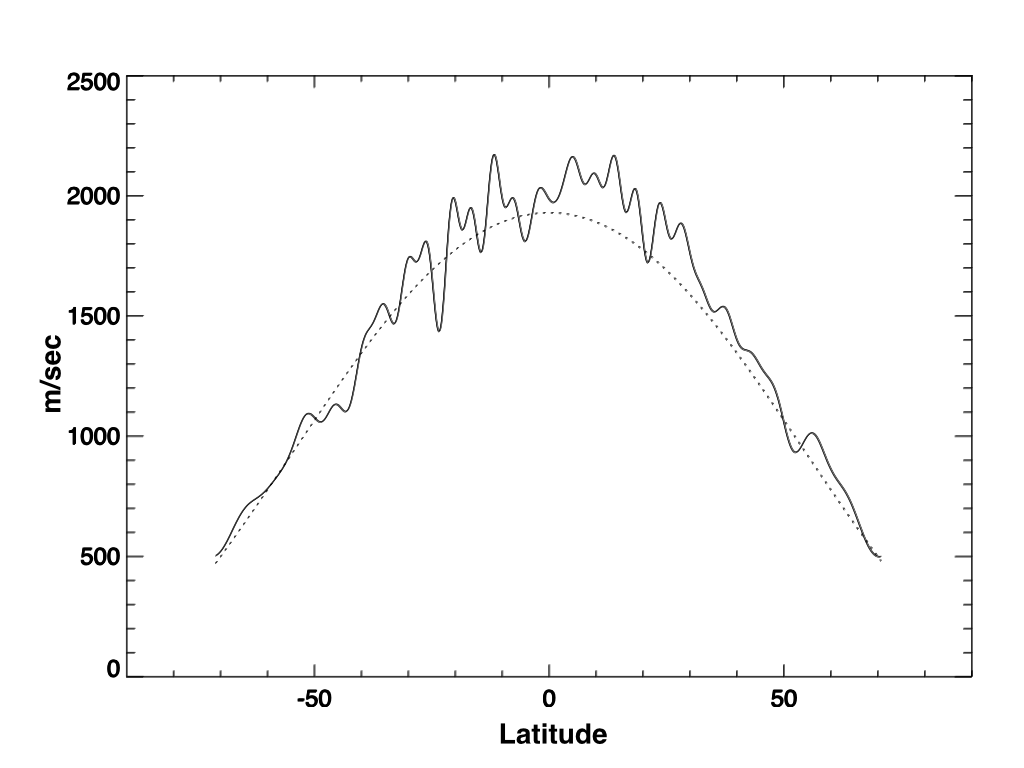}
\caption{The differential rotation of the sun as determined by
applying the spectral optical flow technique on one hour of full-disk
MDI continuum images (solid) and published best fit from Doppler
measurements (dotted). The former was calculated with $N_x$ =4,
$N_y=28$ using sixty $256\times 896$ pixel images. The two agree
within the noise level of the supergranular flow field.}
\label{fig:difrot}
\end{figure}
With this basic validation of the method on perfect data, we turn to comparisons with other methods and operate on real solar data and then consider a detailed error analysis. First we compare the results of applying this method to other large-scale, full-disk measurements.  Figure
\ref{fig:difrot} displays the zonal (E-W) component of the solar velocity along the central meridian of the Sun as a function of solar latitude derived from one hour of MDI \cite{MDI95} data. We include the corresponding measure based upon fits to Doppler measurements \cite{S92}.  Aside from the departures induced by sampling errors of the supergranular flows in this short time, the two curves agree very well. 

\section{Error Assessment}
There are several factors that may contribute to errors in the velocity estimate provided by opflow3d. These can be broken into three classes: systematic errors introduced by the choice of velocity representation, errors due to image quality and errors introduced by physical effects in the solar atmosphere. As a first step, we take what is currently the most consistent and stable images available, using data obtained by the Helioseismic and Magnetic Imager (HMI) on SDO \cite{Scherrer12}. We then subject them to a variety of controlled tests  to address the first two classes of error. This is the best-case scenario for studies of the solar photosphere: a stable imager observing "quiet" sun (where magnetic effects are negligible). The following section considers a more complex situation of observing magnetic regions with a less-stable imager.

One hour of Level-1 HMI continuum images consisting of 80 individual frames were used for this study beginning at 2010-10-26T08:29:00. A set of 1024x1024 pixel sub-images were extracted centered on a coronal hole identified in the Heliophysics Event Knowledgebase \cite{Hurlburt12}  \footnote{SOL2010-10-25T23:00:08L032C113} (herein referred to as C2010).

\begin{figure}
\includegraphics[width=\columnwidth]{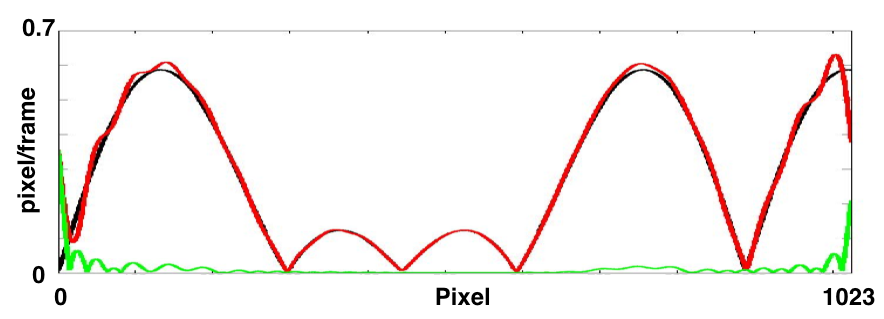}
\caption{The horizontal velocity profile through the middle of the field of view. The magnitude of the simulated flow $|{\bf V}|$ is displayed in black and the fitted velocity $|{\bf v}_f|$, and error, $|{\bf v}-{\bf v}_f|$ in red and green respectively. It is clear that error drops rapidly away from the boundaries, within half a wavelength of the truncated mode.}
\label{fig:Gibbs}
\end{figure}
The use of a truncated Fourier representation for the velocity field is a common practice in fluid dynamic investigations. However it has two well-known issues that must be considered: it imposes a periodic structure to the flows and may introduce aliasing or other errors due to truncation. To assess these effects, we use the first frame from the C2010 and generate ten artificial images from it using equation (\ref{eq:advect}) with a known, hexagonal flow field, in a similar approach to that of Hurlburt \cite{H95}. 

Our implementation uses a fast Fourier transform and can be sensitive to discontinuities at frame boundaries.  While the Fourier representation forces the velocity to match across opposing boundaries, the application of a least squares fit works to confine such effects to near the edges.  The resulting spikes in the residual $|{\bf v}-{\bf v}_f|$  decay rapidly away from frame boundaries (Figure \ref{fig:Gibbs}). Thus, avoiding pixels near the boundaries is the practical means for avoiding these errors.

Aspects of image quality that may influence our fit include large displacements arising from insufficient sampling rates (which may introduce ambiguity into the possible solution), image noise and missing data. To assess the impact of image quality, we adapt the approach from the previous section. The first frame of the C2010 is advected to generate ten frames with a known velocity
field ${\bf V}$ composed of 22 by 22 randomized Fourier modes. A set of these data cubes is generated with a range of magnitudes for ${\bf V}$. We find the accuracy of the fit starts to break down when the RMS velocity exceeds 0.4 pixels/frame.  This result applies to the images with repeating, high-frequency patterns such as solar granulation seen in C2010. In contrast, larger-scale patterns could allow larger unambiguous motions between frames.

The effect of truncation can be assessed by evaluating this same test case with an increasingly large number of modes ($N_x, N_y$) in equation (\ref{eq:vel}), from $N_x, N_y =4$ to $16$. We find that the overall trend is constant, but with higher values being more sensitive to the effects of noise as the effective sample size of the fit decreases. Thus the selection of the number of modes should take this trade off into account. A rule of thumb would be that $N_{x,y} < N_{f},$ where $N_f$ is the number of features (e.g. granules) required to span the image. 

To assess the impact of noise on velocity estimate,  we first measure the inherent noise in the synthetic datacube used above by setting ${\bf v}_f = 0$ in equation (\ref{eq:merit}) to give $\chi(0)  \equiv {\rm RMS}({\partial I \over \partial t}) = 700$ counts per frame.  $\chi(0)$ is used here to represent the original variation that is reduced by fitting V, and because it scales with image contrast. Next, we generate a sequence of datacubes by adding increasing levels of Gaussian white noise. The datacube with added noise standard deviation $\sigma$ has velocity estimate ${\bf v}_f(\sigma)$. By comparing  ${\rm RMS}({\bf v}_f(\sigma)- {\bf V})$  with $ {\rm RMS}({\bf V})$, we can gain some insight into the sensitivity of the method to noise. We find that for a maximum relative error of 1\%, the maximum additional noise must have $\sigma  \le 200$.  Comparing this value of $\sigma$ to the inherent signal noise  $\chi(0)$, demonstrates that high noise images can still yield reliable velocity estimates.  This robustness against measurement noise most likely results from averaging over many pixels.

Finally there are features in the solar atmosphere that may impact the performance of any method of velocity estimation. These include the presence of strong acoustic modes (known as five-minute oscillations) which generate a relatively-smooth, but random intensity fluctuation in solar images; Limb-darkening, which introduces fixed, large-scale intensity gradients due to line-of-sight effects; and strong magnetic features that may distort the intensity patterns in non-obvious ways. In exploring these cases, we cannot compare our results to a known solution. Instead we must make statistical inferences. 

If we seek flows that persist on time scales significantly larger than five minutes, we can assess the effect of acoustic oscillations.  We take ${\bf v}_f(N,j)$ to be the velocity fit for $N$ frames starting at frame $j$.  We examine the convergence of ${\bf v}_f(N,j)$, as $N$ increases.  Using C2010, we divide the 80 frames into sets 1 and 41, consisting of the first and last sets of 40 frames.  Since the images in this case may possess an overall motion akin to camera motion (say ${\bf v}_c(N,j)$), we first subtract such motions from ${\bf v}_f(40,1)$ to produce the the Euclidean metric ${\rm RMS}(|{\bf v}_f(40,1)-{ \bf v}_c(40,1))|) = 0.18$ pixels/frame.  
Similarly, the �distance� between ${\bf v}_f(40,1)$ and ${\bf v}_f(40,41)$ is the Euclidean metric ${\rm RMS}(|{\bf v}_f(40,1)-{\bf v}_f(40,41)|)  =  0.019$ ppf.  This distance is a rough measure of precision.  Thus, with $N = 40$, considerable convergence is apparent.  

To examine the rate of convergence, we can next estimate ${\bf v}_f (2, j)$ for $j=1,39$. We then compute the �distance� ${\rm RMS}({\bf v}_f(2,j) - {\bf v}_f(40,41))$.  This distance, averaged over all $j$, is 0.19 ppf.  Clearly a two frame estimate is poor.  
Comparison of the 2 frame and 40 frame distances, 0.19 ppf and 0.019 ppf, imply that the rate of convergence is between $1/\sqrt{N}$ and $1/N$, which is expected for traveling wave patterns in acoustic oscillations.  This is consistent with previous studies of solar flows using LCT methods \cite{N87}.

The contrast in the image varies smoothly across each frame in our sample due to the limb-darkening effect of the solar atmosphere. Such gradients can cause problems for some methods. However in our case,  it only changes the weights of the sum of the squares in equation (\ref{eq:merit}), so that the fit is only slightly affected.  The same logic shows the method is insensitive to image-quality problems such as missing frames.

\section{Comparison to other methods}
\begin{figure}
\begin{center}
\includegraphics[width=3.5truein]{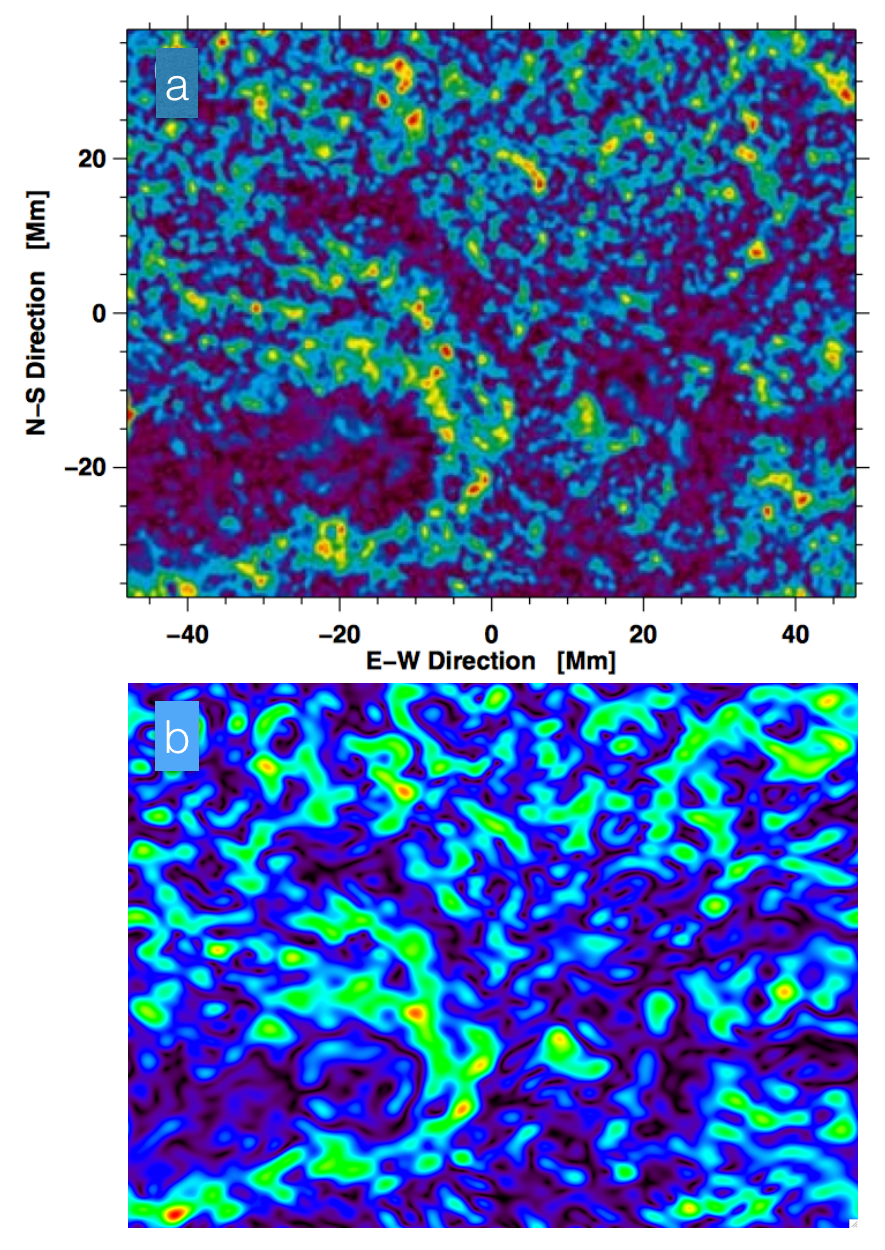}
\end{center}
\caption{A comparison of results of the  method used by (a) V\&D (from their Figure 2c) and (b) {\tt opflow3d} for the same sample of Hinode/SOT data shows detailed agreement. Here we display the magnitude of the two velocity fields using approximately the same color map and scaling from black/dark blue to yellow/red. The flow velocities exhibit the same pattern of outward moat flows around sunspots and inflows around plage. }
\label{fig:Verma2}
\end{figure}

As a final test, we provide a detailed comparison of our method to that used a recent study by Verma \& Decker \cite{Verma11} (hereinafter referred to as V\&D). They conducted a thorough investigation of horizontal flow fields observed in Hinode G-band images using a local correlation tracking (LCT) approach that was used in November and Simon \cite{N87}. 

Following V\&D, we selected an hour-long set of G-band images collected by the Hinode Solar Optical Telescope (SOT, \cite{T08})  on June 4, 2007 between 14:27 and 15:27UT. In that study, the authors first applied the standard calibrations to the images and then further pre-processed them by correcting for foreshortening, applying a rigid alignment between the images to remove spacecraft jitter and solar rotation, and then employed a subsonic filter to remove acoustic oscillations. 

We also calibrate the images using the SolarSoft routine {\tt fg\_prep} with its default settings. However, we do not apply the other preprocessing steps.  Other than foreshortening, those corrections effectively remove noise from the velocity signal that we are seeking, be it jitter from the spacecraft, bulk motion across the field of view or distracting intensity fluctuations. Since the {\tt opflow3d} method has already been shown to address such noise sources, we rely on it alone to do so. In addition, we found seven of the 238 images in the sequence were missing: rather than attempt to correct for these, we left those images blank and left it to {\tt opflow3d} to deal with the consequences.

\begin{figure*}
\begin{center}
\includegraphics[width=\columnwidth]{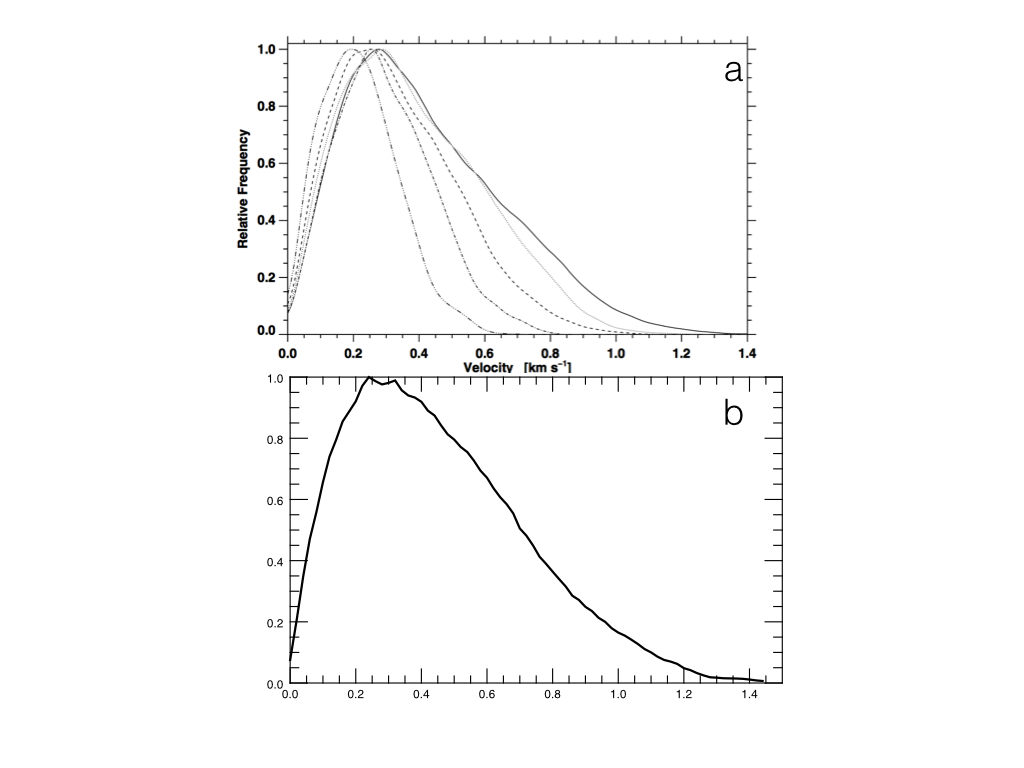}
\end{center}
\caption{Histograms of the speeds found by (a) V\&D (from their Figure 6) and (b) {\tt opflow3d} for the same one-hour sample of Hinode/SOT data also show detailed agreement. The solid curves display the normalized histogram of the velocities for the two methods. (The other curves in (a) correspond to evaluations over longer time intervals from 2 to 16 hours.)}
\label{fig:Verma6}
\end{figure*}

Figures \ref{fig:Verma2} and \ref{fig:Verma6} display a comparison of applying the two methods to the same image set. The only free parameters for {\tt opflow3d} are the number of modes used to fit the velocity field, and whether to use a direct or iterative solution method: we select a direct solution with 20 modes in each direction. This corresponds to an effective pixel size of about 2Mm when compared to the Gaussian FWHM used by V\&D.

We correct for foreshortening after the fact by scaling the components of the velocity and display the magnitude of the resulting velocity field  (less the average velocity over the frame) in Figure \ref{fig:Verma2}. Strong moat flows are visible in the lower left, as well as converging flows elsewhere in plage areas. Figure \ref{fig:Verma6} displays a normalized histogram of flow speeds, which can be compared to figure 6 of V\&D. In both cases the peak value across the field of view is around 0.3 km/s. We find the overall rms speed to be 0.46km/s, the median to be 0.42 km/s and the maximum to be 1.68;  as compared to  0.44 km/s, 0.40 km/s and 1.95 km/s respectively. The fact that {\tt opflow3d} method retains slightly higher rms velocities while reducing the extremes suggests that it might both retain a higher resolving power while mitigating the influences of outliers.

\section{Discussion}
We have described a method for deriving flows from sets of images obtained by a variety of solar imagers. The {\tt opflow3d} procedure has been shown to extract accurate velocity estimates when provided perfect test data and quickly generates results consistent with completely distinct methods when applied on global scales. We have also verified that it agrees in detail with an established method when applied to high-resolution datasets -- and without the need to tune, filter or otherwise preprocess the images before its application. It is currently running as part of the HEK system to identify regions of solar eruptions \cite{Hurlburt15} from data collected by the Atmospheric Imaging Array on SDO \cite{Lemen12}. 

Our method has been found to work well on other types of image data, including magnetograms, since the only assumptions made are that the motions displayed in them are reasonably smooth and persistent. It can also be combined with other image processing methods to extract motions of specific features within the field. For instance the motion of the two polarities (North/South) in magnetograms could be tracked by thresholding the images prior to using {\tt opflow3d}. Similarly, particular scales could be extracted by using high- or low-pass filters.

With the basic approach established, there are several avenues for improvement. First, we could replace the model equation (\ref{eq:advect}) with a more elaborate one, say one that solves the vertical component of the induction equation to extract velocities from sets of magnetograms. Second, we could provide a more elaborate fitting function, say one that permits a simple time dependence. Finally one can seek to optimize the method using more sophisticated tools of linear algebra. We will explore some of these options in future work.

\begin{acknowledgements}
We are grateful to K. Schrijver, R. Shine, T. Tarbell, A. Title and
T. Berger for their helpful conversations and comments. This work has
been supported by the National Aeronautics and Space Administration
through contracts NAS8-39746 and NAS5-3077 and by Lockheed-Martin
Independent Research Funds.
\end{acknowledgements}

\end{document}